\documentclass[runningheads]{llncs}
\usepackage[hidelinks]{hyperref}
\usepackage[T1]{fontenc}

\usepackage{graphicx}
\usepackage{amsmath,amssymb,amsfonts} % math; ok with llncs
\usepackage{mathrsfs}
\usepackage{xcolor}
\usepackage{textcomp}
\usepackage{booktabs}
\usepackage{algorithm}
\usepackage{algpseudocode} % (algorithmicx is loaded by algpseudocode)
\usepackage{listings}
\usepackage{url}
\usepackage{enumitem}
\setlist{nosep}
\usepackage{orcidlink} % Not needed in LLNCS; use \orcidID macro instead

\usepackage{pifont}
\newcommand{\cmark}{\ding{51}}%
\newcommand{\xmark}{\ding{55}}%

\makeatletter
\newcommand{\printfnsymbol}[1]{%
  \textsuperscript{\@fnsymbol{#1}}%
}
\makeatother

\title{Hierarchical Multi-Modal Retrieval for Knowledge-Grounded News Image Captioning}
\titlerunning{Hierarchical Multi-Modal Retrieval for News Image Captioning}

\author{
  Minh-Loi Nguyen\inst{1,2}\orcidlink{0009-0003-2630-3325}\thanks{These authors contributed equally to this work.} \and
  Xuan-Vu Le\inst{1,2}\orcidlink{0009-0009-6094-4912}\printfnsymbol{1} \and
  Long-Bao Nguyen\inst{1,2}\orcidlink{0009-0003-6311-745X} \and
  Hoang-Bach Ngo\inst{1,2}\orcidlink{0009-0002-2290-1187} \and
  Trung-Nghia Le\inst{1,2}\orcidlink{0000-0002-7363-2610}\thanks{Corresponding author.}
}
\authorrunning{M.-L. Nguyen et al.}

\institute{
  University of Science, VNU-HCM, Ho Chi Minh City, Vietnam \and
  Vietnam National University, Ho Chi Minh City, Vietnam \\
  \email{\{22120189, 22120438, 22120025\}@student.hcmus.edu.vn, nhbach22@apcs.fitus.edu.vn, ltnghia@fit.hcmus.edu.vn}
}

\begin{document}
\maketitle

\begin{abstract}
Traditional image captioning methods often struggle to generate comprehensive, context-rich descriptions, especially for details not directly observable from visual cues. To overcome this, we propose a novel retrieval-augmented image captioning framework that generates captions with deeper insights, such as object attributes, event context, and underlying significance, by leveraging external knowledge. Our approach features a hierarchical multi-modal article retrieval mechanism that moves beyond monolithic text entities. This retrieval considers article structure-aware features, including weighted textual components (e.g., headlines, body sections) and visual placement patterns, alongside multi-faceted similarity computations (content--visual, visual--visual, and discourse positioning). A subsequent contextual relevance refinement stage further enhances the retrieved information. The retrieved articles then serve as the knowledge base for caption generation: first, a VLM generates a concise image description; second, we segment relevant information from the retrieved articles based on this description; and finally, an LLM utilizes both the description and extracted knowledge to generate a comprehensive, contextually detailed caption. We participated in the ACM Multimedia EVENTA 2025 Challenge and achieved 5th place with an overall score of 0.2824 on the private test set of the OpenEvent-V1 dataset. Source code is publicly released at \url{https://github.com/mf0212/EVENTA-Challange}.
\end{abstract}


\section{Introduction}\label{sec:intro}

% \comment{sua loi citation. dung bib file, khong nhung truc tiep vao tex file.}

Image captioning has emerged as a pivotal task in computer vision and natural language processing, aiming to generate textual descriptions that accurately reflect the content of an image. While these traditional methods excel at describing visible elements, such methods frequently fall short in providing comprehensive, context-rich information, particularly regarding novel objects or deeper insights beyond what is immediately observable in the image. This limitation highlights a critical need for approaches that can enrich captions with richer details, such as object attributes, event timing, contextual relevance, and underlying significance, which are not directly observable from visual cues alone.

To address these challenges, retrieval-augmented image captioning has emerged as a promising paradigm, building upon foundational work in information retrieval and generative models. Pioneering efforts such as Retrieval-Augmented Generation (RAG) \cite{lewis2020retrieval} demonstrated the efficacy of combining dense vector retrieval with generative models to retrieve relevant documents, thereby providing a robust framework for integrating external knowledge into generative tasks. In the realm of image captioning, this approach leverages external knowledge to enhance caption generation by incorporating retrieved textual evidence. Sarto et al. \cite{sarto2022retrieval} proposed a retrieval-augmented transformer model that uses a kNN memory to retrieve relevant captions based on visual similarities, significantly improving caption quality on the COCO dataset \cite{lin2014microsoft}. EXTRA model \cite{ramos2023extra} retrieves captions from a datastore (e.g., training captions or external ones), encodes them together with the image via a Vision-\&-Language BERT \cite{devlin-etal-2019-bert}, and then decodes a caption conditioned on both the image and the retrieved textual context. Using enough retrieved captions (k=5) improves performance. EVCAP \cite{li2024evcap} advanced this field by utilizing an external visual-name memory to prompt vision language models (VLMs) \cite{ijcai2022p762} with retrieved object names, thereby facilitating open-world comprehension and reducing reliance on extensive training data or large network parameters \cite{li2024evcap}. Additionally, Wu et al. \cite{wu2024dir} enhanced image-to-text retrieval using diffusion-guided feature learning and a high-quality retrieval database, improving out-of-domain generalization.

% Our research aims to further advance this field by generating captions that offer richer, more comprehensive information about an image, particularly in scenarios where contextual details are paramount, such as news-related applications. These captions extend beyond simple visual descriptions to provide deeper insights, including the names and attributes of objects, the timing, context, and outcomes of events, and other crucial details that cannot be gleaned from merely observing the image. To achieve this goal, our approach employs a sophisticated hierarchical multi-modal framework that considers articles not as monolithic entities, but analyzes their internal structure, including headlines, body sections, and image captions, alongside visual placement patterns. This enables multi-faceted similarity computation and contextual relevance refinement for robust article selection. To this end, given an image, our method first performs a sophisticated retrieval of relevant articles from a provided external database, a crucial step enabling the extraction of necessary information to generate an enriched image caption. \comment{viet them vai cau nua de neu chi tiet hon mot chut ve ky thuat cua phan nay}
Our research aims to further advance this field by generating captions that offer richer, more comprehensive information about an image, particularly in scenarios where contextual details are paramount, such as news-related applications. These captions extend beyond simple visual descriptions to provide deeper insights, including the names and attributes of objects, the timing, context, and outcomes of events, and other crucial details that cannot be gleaned from merely observing the image. To achieve this goal, our approach employs a hierarchical multi-modal framework that analyzes article structure (headlines, body sections, image captions) alongside visual placement patterns for multi-faceted similarity computation and robust article selection.

Our proposed captioning module then integrates this retrieved knowledge with image understanding through a three-step process. Initially, a Vision Language Model (VLM) is employed to generate a concise description of the image, focusing on its main theme, composition, and prominent objects; this initial description serves as a crucial bridge between the visual content and the external knowledge base. Subsequently, based on this concise description, relevant information is segmented from the retrieved articles, thereby extracting useful knowledge pertinent to the initial description, a step inspired by approaches like that emphasize targeted information retrieval to augment VLM capabilities \cite{li2024understanding}. Ultimately, both the concise image description and the extracted relevant information from the article are provided to LLM to generate the comprehensive, final caption.

% \comment{We participated in the ACM Multimedia EVENTA 2025 Challenge \cite{eventa25} and achieved 5th place with an overall score of 0.2824 on the private test set of the OpenEvent-V1 dataset \cite{nguyen2025openevents}. This result highlights ....... Our contributions are as follows:}
% \begin{itemize}
%     \item \comment{We present a novel hierarchical multi-modal retrieval-augmented image captioning framework, which ......}
%     \item \comment{We introduce a hierarchical multi-modal article retrieval to .....}
%     \item \comment{We propose a novel two-stage context-aware LLM-driven image captioning to synthesize rich visual information with verified textual knowledge. }
%     \item \comment{We achieved 5th place in the ACM Multimedia EVENTA 2025 Challenge.}
% \end{itemize}
We participated in the ACM Multimedia EVENTA 2025 Challenge \cite{eventa25} and achieved 5th place with an overall score of 0.2824 on the private test set of the OpenEvent-V1 dataset \cite{nguyen2025openevents}. This result highlights the effectiveness of our hierarchical approach in real-world news captioning scenarios. 

Our contributions are as follows:
\begin{itemize}
    \item We present a hierarchical multi-modal retrieval-augmented captioning framework that decomposes news articles into structured components with differential weighting. Source code: \url{https://github.com/mf0212/EVENTA-Challange}.
    \item We design a retrieval system that combines content-visual alignment, visual-visual coherence, and discourse positioning to identify relevant information.
    \item We propose a three-stage LLM-driven captioning pipeline that integrates visual context, relevant sentences, and knowledge-grounded synthesis.
    \item Our system ranked 5th in the ACM Multimedia EVENTA 2025 Challenge with a score of 0.2824.
\end{itemize}

\section{Related Work}\label{sec:related}

% \subsection{Information Retrieval}

\textbf{Information retrieval} is essential for fetching articles that align with image content. Lewis et al. \cite{lewis2020retrieval} introduced RAG, combining dense vector retrieval with generative models. Ji et al. \cite{ji2023multimodal} leveraged Wikipedia for multi-modal article retrieval. Yang et al. \cite{yang2023semantic} proposed semantic search using transformers in shared latent spaces. Nguyen et al. \cite{nguyen2023multimodal} explored multi-modal fusion for news article matching using CLIP and keyphrase extraction. Ngo et al. \cite{ngo2023comprehensive} advanced cross-modal retrieval for vehicle detection. Our method builds on these by using image-derived captions as queries, optimizing for news contexts with high visual-textual alignment.

% \subsection{Augmented Image Captioning}

\textbf{Augmented image captioning} incorporates external knowledge for contextually rich descriptions. Ramos et al. \cite{ramos2023extra} proposed EXTRA using BERT \cite{devlin-etal-2019-bert} with retrieved captions. Sarto et al. \cite{sarto2022retrieval} introduced kNN memory-augmented transformers. Li et al. \cite{li2024understanding} prompted LLMs with retrieved object names. Wu et al. \cite{wu2024dir} employed diffusion-guided retrieval. Our method generates LLM-based descriptions, segments relevant article information, and synthesizes both for comprehensive, contextually detailed captions tailored for news applications.

\section{Proposed Method}\label{sec:methods}

\subsection{Overview}

\begin{figure}[t!]
\centering
\includegraphics[width=\textwidth]{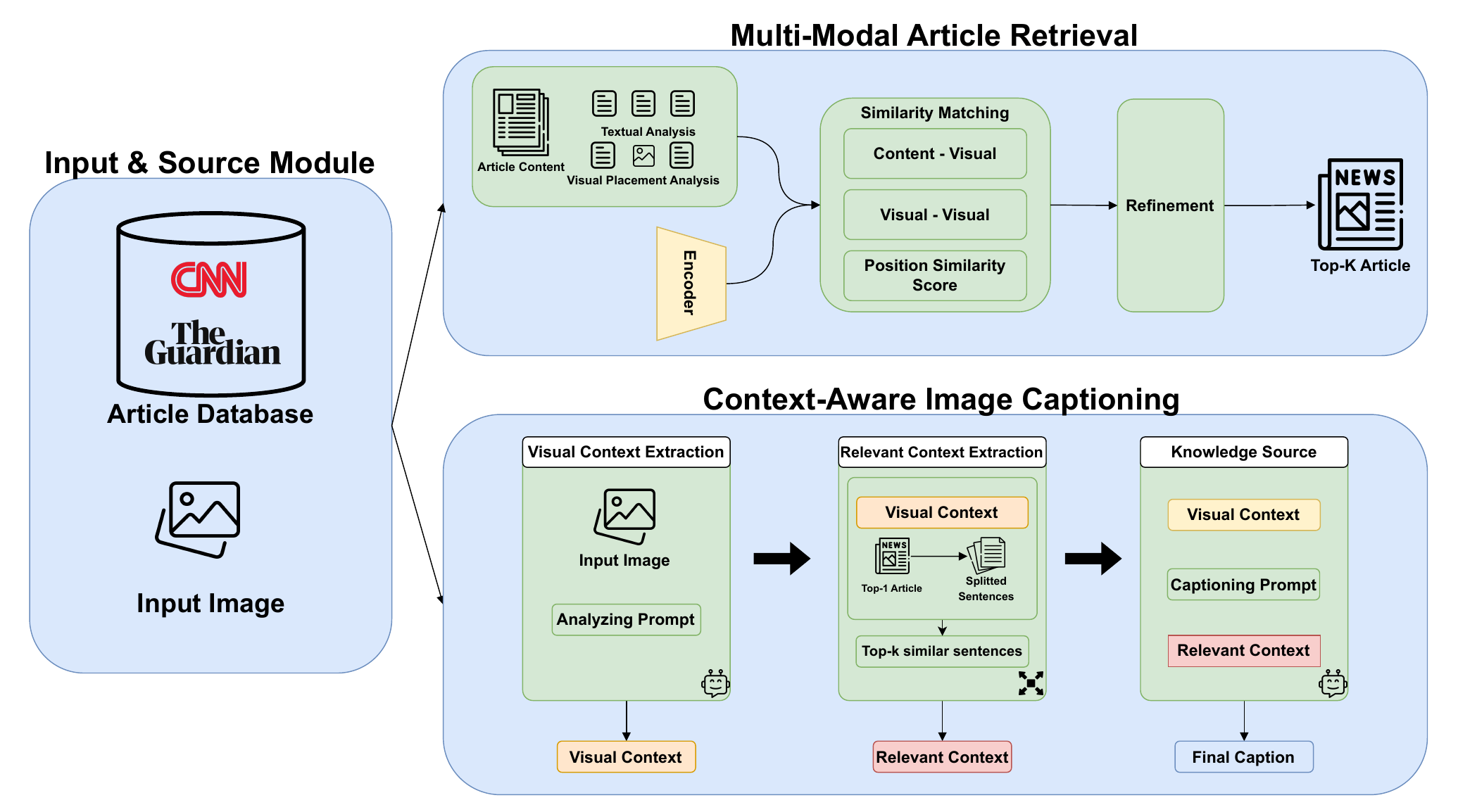}
\caption{Overview of our hierarchical multi-modal retrieval-augmented image captioning framework. The system operates in two main phases: (1) Multi-Modal Article Retrieval - Given an input image, the system retrieves contextually relevant articles from a structured database (CNN/The Guardian) using our novel multi-faceted similarity computation that combines content-visual alignment, visual-visual coherence, and discourse positioning scores; (2) Context-Aware Image Captioning - A three-stage pipeline that first extracts structured visual context from the input image, then identifies and extracts the most relevant sentences from the top-retrieved article, and finally synthesizes a comprehensive, knowledge-grounded caption using a large language model that integrates both visual understanding and factual article content.}
\label{fig:overview}
\end{figure}

% \comment{viet gioi thieu ngan gon ve proposed framework o fig \ref{fig:overview}. bao gom 2 phan chinh: retrieval de lam gi; va captioning de lam gi}

Our proposed framework, illustrated in Fig. \ref{fig:overview}, consists of two main components working in tandem. The Multi-Modal Article Retrieval module leverages the inherent structure of news articles to identify the most contextually relevant documents from a large-scale database, moving beyond simple text matching to consider visual coherence and discourse patterns. The Context-Aware Image Captioning module then synthesizes the retrieved knowledge with visual understanding through a carefully orchestrated three-stage process, ensuring that the final captions are both visually grounded and factually enriched with specific details such as names, dates, and event significance.

% \subsection{Hierarchical Multi-Modal Article Retrieval}

% Rather than treating articles as monolithic text entities, our approach recognizes that news articles exhibit rich internal structure that provides crucial semantic signals for event-image alignment. We propose a hierarchical retrieval framework that jointly considers textual semantics, visual placement patterns, and structural discourse elements within articles to identify contextually relevant information sources.

% \comment{noi so luoc ve pipeline, gom co abc xyz. muc dich paragraph nay de lien ket cac subsubsection}

\subsection{Hierarchical Multi-Modal Article Retrieval}

Rather than treating articles as monolithic text entities, our approach recognizes that news articles exhibit rich internal structure that provides crucial semantic signals for event-image alignment. We propose a hierarchical retrieval framework that jointly considers textual semantics, visual placement patterns, and structural discourse elements within articles to identify contextually relevant information sources.

Our retrieval pipeline consists of three core components: (i) Article Structure-Aware Feature Extraction that decomposes articles into weighted textual components and placement-aware visual features, (ii) Multi-Faceted Similarity Computation that evaluates query-article relevance through content, visual, and positional channels, and (iii) Contextual Relevance Refinement that leverages temporal and citation relationships to enhance retrieval accuracy. This hierarchical approach ensures that retrieved articles are not only semantically relevant but also structurally aligned with the query image's journalistic context.

\subsubsection{Article Structure-Aware Feature Extraction}

\textbf{Spatial-Semantic Text Analysis.} We decompose each article $a_i$ into structurally meaningful components: headline $H_i$, lead paragraph $L_i$, body sections $B_i = \{b_1, b_2, ..., b_k\}$, and image captions $C_i = \{c_1, c_2, ..., c_m\}$. Each component receives differential weighting based on its journalistic significance:

\begin{equation}
\mathbf{t}_{a_i} = \alpha_H f_t(H_i) + \alpha_L f_t(L_i) + \alpha_B \frac{1}{|B_i|}\sum_{j=1}^{|B_i|} f_t(b_j) + \alpha_C \frac{1}{|C_i|}\sum_{j=1}^{|C_i|} f_t(c_j),
\end{equation}
where $\alpha_H$, $\alpha_L$, $\alpha_B$, and $\alpha_C$ represent learned importance weights for headlines, leads, body text, and captions, respectively, and $f_t(\cdot)$ denotes the text encoding function implemented using a pre-trained language model (e.g., BERT or sentence transformers) that maps textual content to dense vector representations.

\textbf{Visual Placement Feature Integration.} Beyond textual content, we extract placement-aware visual features from images embedded within articles. The hypothesis is that image positioning and co-occurrence patterns with specific text segments provide semantic cues about event relevance. For each article image $I_{a,j}$ positioned at text location $\ell_j$, we compute $\mathbf{p}_{a,j} = f_v(I_{a,j}) \odot g(\ell_j, \text{context}_{a,j}),$ where $f_v(\cdot)$ represents the visual encoding function implemented using CLIP-ViT-B/32, which extracts high-dimensional feature representations from images, $g(\cdot)$ is a learned multi-layer perceptron that maps position indices (normalized by article length) and surrounding text context embeddings to a modulation vector, and $\odot$ denotes element-wise modulation: $\mathbf{v}_{a_i} = \frac{1}{|I_a|}\sum_{j=1}^{|I_a|} \mathbf{p}_{a,j}.$

\subsubsection{Multi-Faceted Similarity Computation}

Our retrieval mechanism operates through three complementary similarity channels that capture different aspects of event-article relevance: we compute the \textit{Content-Visual Alignment} as primary similarity between query image $q$ and article text content $a_i$: $s_{\text{content}}(q, a_i) = \cos(\mathbf{v}_q, \mathbf{t}_{a_i}).$ For \textit{Visual-Visual Coherence}, we compute cross-image similarity between query and article-embedded images $(q, a_i)$: $s_{\text{visual}}(q, a_i) = \max_{j \in |I_a|} \cos(\mathbf{v}_q, \mathbf{p}_{a,j}).$ For \textit{Discourse Positioning Score}, we introduce a novel metric capturing how image placement within article structure correlates with event centrality: $s_{\text{position}}(q, a_i) = \sum_{j=1}^{|I_a|} w(\ell_j) \cdot \cos(\mathbf{v}_q, \mathbf{p}_{a,j}),$ where $w(\ell_j)$ assigns higher weights to images positioned near crucial discourse markers (e.g., headlines, first paragraphs, conclusion sections). The composite retrieval score integrates these channels through learned combination:
\begin{equation}
\text{score}(q, a_i) = \beta_1 s_{\text{content}} + \beta_2 s_{\text{visual}} + \beta_3 s_{\text{position}}.
\end{equation}

% \subsubsection{Contextual Relevance Refinement}

% Beyond similarity computation, we implement a contextual refinement stage that leverages cross-article relationships. Articles frequently reference related events, and we exploit these connections through:

% \textbf{Temporal Clustering.} Articles published within temporal windows often cover related event aspects. We group articles by publication timestamps and boost scores for articles from clusters containing high-similarity items.

% \textbf{Citation Network Analysis.} When available \comment{what do you mean?}, we analyze inter-article references and boost retrieval scores for articles that are frequently co-referenced with initially high-scoring items.

% \comment{phan nay con kha so sai, kho hieu. mo ta chi tiet them.}

\subsubsection{Contextual Relevance Refinement}

Beyond similarity computation, we implement a contextual refinement stage that leverages cross-article relationships. Articles frequently reference related events, and we exploit these connections through:

\textbf{Temporal Clustering.} We group articles by publication timestamps using a sliding window approach (window size = 7 days). For each retrieved article $a_i$ with timestamp $t_i$, we identify its temporal cluster $C_t(a_i) = \{a_j : |t_j - t_i| < \delta_t\}$. The refined score incorporates cluster coherence:
\begin{equation}
\text{score}_{\text{temporal}}(q, a_i) = \text{score}(q, a_i) + \gamma \cdot \frac{1}{|C_t(a_i)|} \sum_{a_j \in C_t(a_i)} \text{score}(q, a_j)
\end{equation}
where $\gamma$ is a weighting parameter (set to 0.2 in our experiments).

\textbf{Citation Network Analysis.} When citation metadata is available in the dataset (e.g., hyperlinks between articles or explicit references), we construct a directed graph $G = (V, E)$ where nodes represent articles and edges represent citations. We boost retrieval scores for articles that share citation neighborhoods with high-scoring items:
\begin{equation}
\text{score}_{\text{final}}(q, a_i) = \text{score}_{\text{temporal}}(q, a_i) \cdot (1 + \lambda \cdot \text{PageRank}(a_i, G))
\end{equation}
where PageRank scores are pre-computed on the citation graph and $\lambda$ controls the influence of citation importance (set to 0.1 in our experiments).

\subsubsection{Implementations}

We implemented the retrieval module using CLIP-ViT-B/32 \cite{radford2021learning} as the base encoder, with additional learned projection layers for position encoding and component weighting. The hierarchical weights $\{\alpha_H, \alpha_L, \alpha_B, \alpha_C\}$ and combination parameters $\{\beta_1, \beta_2, \beta_3\}$ were optimized through validation on a held-out subset of the dataset.

\subsection{Context-Aware Image Captioning}

Following the retrieval of contextually relevant articles, our system proceeds to the caption generation phase. We eschew a simple image-to-text approach, which often fails to capture the specific nuances and factual details of news events. Instead, we propose a novel two-stage, LLM-driven captioning pipeline designed to synthesize rich visual information with verified textual knowledge. This process decomposes the complex task of news captioning into three manageable steps. %: (1) structured visual context extraction, (2) relevant context extraction, and (3) knowledge-grounded caption synthesis.

\subsubsection{Stage 1: Structured Visual Context Extraction}

The first stage aims to produce a comprehensive, multi-faceted analysis of the input image ($I_q$) in isolation. This serves as a structured intermediate representation that forces the model to deeply "understand" the visual evidence before being exposed to external knowledge, thereby mitigating the risk of the textual context overpowering the visual content. We employ a Vision-Language Model (VLM) prompted to act as a "visual analyst." The model is instructed to generate a detailed description, $\mathcal{D}_{\text{visual}}$, by addressing four key analytical dimensions:

\begin{itemize}
    \item \textbf{Objective Description:} A factual account of the scene, including subjects, objects, setting, and any legible text.
    \item \textbf{Contextual Inference:} A preliminary hypothesis about the event, its location, and subject matter, based purely on visual cues.
    \item \textbf{Mood \& Atmosphere:} An assessment of the emotional tone and overall atmosphere conveyed by the image (e.g., celebratory, tense, somber).
    \item \textbf{Potential Headline:} A speculative news headline that encapsulates the core visual narrative.
\end{itemize}

The output of this stage is a single, cohesive paragraph that integrates these four facets, forming a rich descriptive prior, $\mathcal{D}_{\text{visual}}$, for the subsequent synthesis stage.

\subsubsection{Stage 2: Relevant Context Extraction}

\begin{figure}[t!]
\centering
\includegraphics[width=\textwidth]{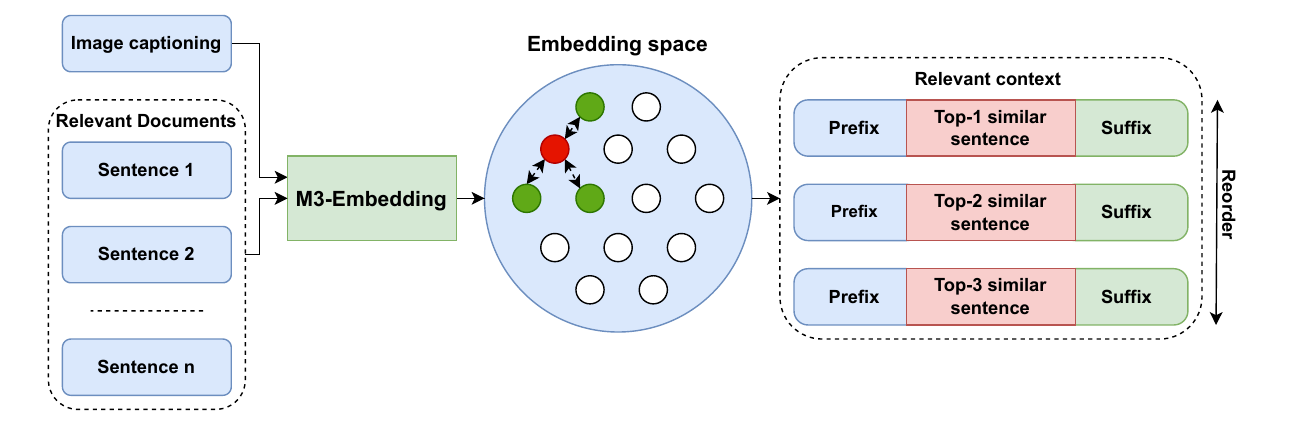}
%\vspace{-8mm}
\caption{Overview of the relevant context extraction stage. Relevant documents are split into sentences before embedding, and the final relevant context is constructed by including the prefix and suffix around the top-3 most similar sentences.}
\label{fig:relevant}
\end{figure}

 Fig. \ref{fig:relevant} describe this stage, which begins by encoding the structured visual context $\mathcal{D}_{\text{visual}}$ using the M3-embedding \cite{muennighoff2024m3}, a model for language embedding, rather than relying on multimodal embeddings. This decision is driven by two key observations:
\begin{enumerate}
    \item Language-only embeddings consistently outperform multimodal representations in our retrieval setting.
    \item Our image descriptions achieve high \textbf{CLIPScore}, indicating strong semantic alignment with the image content.
\end{enumerate}

We rely on the top-1 retrieved document, as including multiple sources often failed to capture key details and introduced noise. Using a single, highly relevant document ensures the model remains focused on the most informative content. The document is segmented into sentences with a regex-based splitter, preserving semantic completeness compared to arbitrary or fixed-length chunking.

Each sentence is embedded with the same M3-embedding model used for the query, and cosine similarity scores are computed against the query embedding. Sentences are then ranked, and the top-3 are selected as the most relevant evidence. To provide richer context, each selected sentence is expanded with up to two neighboring sentences from the original document, and these chunks are reordered to maintain coherence. The aggregated content $\mathcal{A}_{\text{retrieved}}$ is then passed to the LLM, allowing it to capture both precise evidence and local context, resulting in more accurate and coherent captions.

\subsubsection{Stage 3: Knowledge-Grounded Caption Synthesis}

The third stage is responsible for the final caption generation. Its core function is to skillfully weave together the structured visual analysis ($\mathcal{D}_{\text{visual}}$) with the factual context provided by the top-$k$ retrieved articles, $\mathcal{A}_{\text{retrieved}}$, from our hierarchical retrieval module.

We utilize a powerful LLM configured as an "expert news photo caption writer." This model receives a carefully engineered prompt that mandates a synthesis-driven methodology. The key inputs to this prompt are:
\begin{enumerate}
    \item The input image $I_q$.
    \item The structured visual context $\mathcal{D}_{\text{visual}}$ generated in Stage 1.
    \item The aggregated text content from the retrieved articles $\mathcal{A}_{\text{retrieved}}$ in Stage 2.
\end{enumerate}

The prompt explicitly instructs the model to anchor its narrative in the visual evidence described in $\mathcal{D}_{\text{visual}}$ while using the article content $\mathcal{A}_{\text{retrieved}}$ to inject specific, factual details such as names of individuals, precise locations, dates, and the underlying cause or significance of the event. This ensures the final caption, $C_{\text{final}}$, is not merely a summary of the articles but a direct, context-rich explanation of the moment captured in the photograph.

This three-stage design provides significant advantages. The decomposition of tasks ensures that both visual fidelity and contextual accuracy are maintained. Furthermore, the intermediate representation $\mathcal{D}_{\text{visual}}$ offers a degree of interpretability and allows the pipeline to ground the final narrative firmly in what is actually visible, producing captions that are both informative and faithful to the source image.

\section{Experiments}\label{sec:experiments}

% In this section, we detail the experiments conducted to validate our proposed pipeline. We begin by outlining the experimental setup, including the dataset, evaluation metrics, and baselines. We then present a series of ablation studies to analyze the contribution of each component in our retrieval and captioning modules. Finally, we report our system's official performance on the competition's private leaderboard and provide a qualitative analysis to illustrate its capabilities.

\subsection{Experimental Settings}

% \subsubsection{Dataset}
All experiments were performed on the OpenEvents V1 dataset \cite{nguyen2025openevents}, the official EVENTA 2025 Challenge dataset \cite{eventa25}, comprising 200,000+ news articles and 400,000+ images from CNN and The Guardian (2011–2022).

% \subsubsection{Evaluation Metrics}
Following the official challenge guidelines \cite{eventa25}, we evaluated our system using Mean Average Precision (\textbf{mAP}), Recall@1 (\textbf{R@1}), and Recall@10 (\textbf{R@10}) for retrieval evaluation, \textbf{CIDEr} and \textbf{CLIP Score} \cite{hessel2021clipscore} for captioning evaluation, and the \textbf{Overall Score} defined by the challenge organizers to rank participants based on performance across both tasks.

\subsection{Ablation Study}

% \subsubsection{Baselines}
% To demonstrate the effectiveness of our proposed pipeline, we established several baselines for comparison in our ablation studies:
% \begin{itemize}
%     \item \textbf{CLIP-based Retrieval:} A standard baseline that computes cosine similarity between the CLIP embedding of the query image and the CLIP embedding of the entire raw text of each article.
%     \item \textbf{Image-to-Text (No Retrieval):} A standard Vision-Language Model (VLM) that generates a caption from the input image alone, without any retrieved article context.
%     \item \textbf{One-Stage Captioning:} A VLM that receives both the image and the raw text of the best-retrieved article as input, generating a caption in a single step without the structured visual analysis stage.
% \end{itemize}

\subsubsection{Analysis of Retrieval Module}

\begin{table}[t!]
\centering
\caption{Effectiveness of different features in our the retrieval module on the public test set. The full model demonstrates superior performance, confirming the benefits of combining content, visual, and positional signals.}\label{tab:retrieval_ablation}
\begin{tabular}{@{}lccccccc@{}}
\toprule
\textbf{Method} & CLIP & Content & Visual & Position & \textbf{mAP} & \textbf{R@1} & \textbf{R@10} \\
\midrule
Baseline 1 & \cmark & \xmark & \xmark & \xmark & 0.11 & 0.08 & 0.10 \\
% \midrule
Baseline 2 & \cmark & \cmark & \xmark & \xmark & 0.15 & 0.12 & 0.16 \\
Baseline 3 & \cmark & \cmark & \cmark & \xmark & 0.95 & 0.942 & 0.985 \\
\textbf{Ours} & \cmark & \cmark & \cmark & \cmark & \textbf{0.97} & \textbf{0.956} & \textbf{0.991} \\
\bottomrule
\end{tabular}
% %\vspace{-7mm}
\end{table}

To validate our hierarchical multi-modal retrieval framework, we conducted systematic ablation studies on the public test set, examining the individual and collective contributions of each similarity computation channel. Table~\ref{tab:retrieval_ablation} presents the progressive performance improvements achieved through our multi-faceted approach. The results demonstrate several key findings. First, our structure-aware content similarity ($s_{\text{content}}$) alone outperforms the CLIP baseline across all metrics, validating the importance of decomposing articles into semantically meaningful components with differential weighting. Second, incorporating visual-visual coherence ($s_{\text{visual}}$) yields dramatic improvements, with mAP increasing from 0.15 to 0.95, confirming that cross-image similarity between query and article-embedded images serves as a highly discriminative relevance signal. Finally, the full model integrating our discourse positioning score ($s_{\text{position}}$) achieves optimal performance (mAP: 0.97, R@1: 0.956, R@10: 0.991), validating our hypothesis that image placement within news article discourse structure correlates meaningfully with event centrality. These results establish that multi-modal retrieval systems benefit significantly from the synergistic integration of structural, visual, and positional signals rather than relying solely on semantic similarity measures.

\subsubsection{Analysis of Captioning Module}

\begin{table}[t!]
\centering
\caption{Performance comparison of captioning methods on the public test set. For the baseline, that generates a caption from the input image alone, without any retrieved article context. Our three-stage pipeline, which ..., significantly outperforms the baselines.}\label{tab:captioning_ablation}
\begin{tabular}{@{}lcc@{}}
\toprule
\textbf{Method} & \textbf{CIDEr} & \textbf{CLIP Score} \\
\midrule
Baseline (Image-to-Text without Retrieval) & 0.039 & 0.890 \\
\textbf{Ours (Three-Stage)} & \textbf{0.123} & \textbf{0.883} \\
\bottomrule
\end{tabular}
%\vspace{-7mm}
\end{table}

We evaluated our proposed two-stage captioning pipeline against the baselines on the public test set. As shown in Table~\ref{tab:captioning_ablation}, the \textit{Image-to-Text} model yields poor scores, producing generic captions that lack factual grounding. Our \textbf{Three-Stage Pipeline} achieves the highest scores. This confirms the efficacy of our design: the first stage generates a structured visual analysis that "anchors" the model to the image content, preventing the LLM in the second stage from merely summarizing the text. This separation of concerns leads to captions that are both visually faithful and contextually rich.

\subsection{Overall Performance}

\begin{table}[t!]
\centering
\caption{Official results on the EVENTA Grand Challenge private test set leaderboard. Our entry is "noname\_".}\label{tab:leaderboard_results}
\begin{tabular}{@{}clcccccc@{}}
\toprule
\textbf{\#} & \textbf{Participant} & \textbf{Overall Score} & \textbf{mAP} & \textbf{R@1} & \textbf{R@10} & \textbf{CLIP Score} & \textbf{CIDEr} \\
\midrule
1 & cerebro & \textbf{0.5501} & 0.991 & 0.989 & 0.995 & 0.826 & \textbf{0.210} \\
2 & SodaBread & 0.5457 & 0.982 & 0.977 & 0.988 & \textbf{0.870} & 0.204 \\
3 & Re: Zero Slavery & 0.4515 & 0.955 & 0.945 & 0.973 & 0.732 & 0.156 \\
4 & ITxTK9 & 0.4200 & 0.966 & 0.955 & 0.983 & 0.828 & 0.133 \\
\textbf{5} & \textbf{noname\_ (Ours)} & 0.2824 & 0.708 & 0.663 & 0.801 & 0.783 & 0.081 \\
\bottomrule
\end{tabular}
%\vspace{-5mm}
\end{table}

Table~\ref{tab:leaderboard_results} presents the official results of our system (team name: \textbf{noname\_}) on the private test set leaderboard, benchmarked against the top-performing teams. Our system achieved an overall score of 0.2824, placing 5th in the challenge. An analysis of the individual metrics provides further insight. Our retrieval scores (mAP 0.708, R@1 0.663) demonstrate the fundamental effectiveness of our method, though a performance gap exists compared to the top-ranked systems. This suggests that the leading methods may have utilized more extensive model fine-tuning or sophisticated data augmentation techniques. In the captioning task, our CIDEr score of 0.081 reflects the immense difficulty of generating news-style captions that precisely match human-written references, a known challenge for metrics sensitive to n-gram overlap. However, our strong CLIP Score (0.783) indicates that our generated captions are semantically highly relevant to the images, validating the core principle of our synthesis module. This discrepancy suggests that while our captions capture the correct meaning, they may differ lexically from the ground truth.

\begin{figure}[t!]
    \centering
    \includegraphics[width=\linewidth]{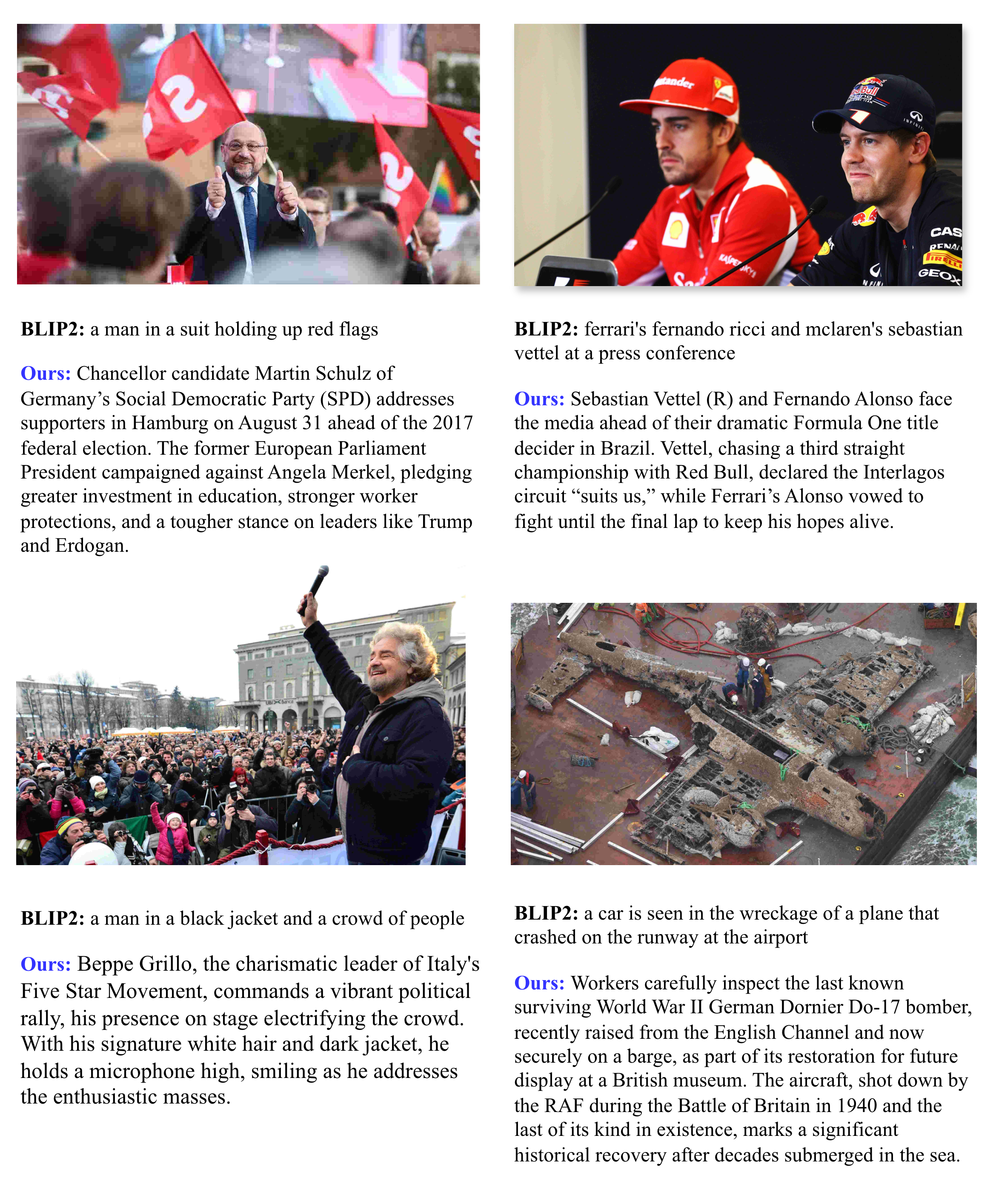}
    \vspace{-8mm}
    \caption{Qualitative examples comparing baseline captions (top, black text) against our pipeline (bottom, blue text). Unlike generic baseline descriptions, our method generates context-aware, factually accurate captions by leveraging retrieved articles.}
    \label{fig:qualitative}
\end{figure}

To better illustrate the practical value of our approach, a qualitative comparison (Figure \ref{fig:qualitative}) shows that a baseline captioner might describe the left image simply as “a man in a suit smiling and gesturing with his thumbs up in front of red flags” or the right image as “two men wearing caps sitting at microphones.” In contrast, our pipeline, by leveraging the retrieved article, correctly identifies the individuals, the context of the event, and its significance. For example, it can generate a caption such as “Martin Schulz, leader of the Social Democratic Party, campaigns at a rally in Berlin ahead of the federal elections” or “Fernando Alonso and Sebastian Vettel address the media during a Formula 1 press conference.” This pattern also appears in the other two examples: while a baseline method would only say “a man in a black jacket speaking to a crowd,” our system can recognize Beppe Grillo and describe the political rally he is leading, and instead of outputting a vague description such as “a damaged vehicle in debris,” it can correctly identify the scene as workers examining the recovered remains of a World War II Dornier Do-17 bomber. Across all four cases, the model moves beyond generic object-level descriptions and instead produces grounded, event-aware captions that capture who is involved, what is happening, and why the image matters. While quantitative metrics provide a broad view of system performance, these qualitative examples reveal strengths that traditional captioning scores may fail to capture, particularly when correctness depends on integrating external knowledge rather than visual similarity. This example demonstrates that despite the quantitative gap in some metrics, our pipeline successfully achieves its primary goal: producing visually grounded, context-aware, and factually accurate news image captions.

\section{Conclusion}\label{sec:conclusion}

We presented a hierarchical multi-modal retrieval-augmented framework for knowledge-grounded news image captioning that addresses the limitations of traditional methods in providing comprehensive, context-rich descriptions. Our approach treats news articles as structured entities with differential component weighting, achieving significant improvements in both retrieval effectiveness and caption quality. Our framework transforms generic visual descriptions into specific, informative narratives enriched with names, dates, locations, and event significance, highlighting its value for news, accessibility, and automated journalism. Future work includes extending to other document types, integrating additional relevance signals, and improving efficiency via parallel generation.

% ---- Credits / Acknowledgments (LLNCS 2.21 style) ----
\begin{credits}
\subsubsection{\ackname}
This research is supported by research funding from Faculty of Information Technology, University of Science, VNU-HCM. This research used the GPUs provided by the Intelligent Systems Lab at the Faculty of Information Technology, University of Science, VNU-HCM.
\end{credits}

% ---- Bibliography ----
\bibliographystyle{splncs04}
\bibliography{ref}

@inproceedings{lewis2020retrieval,
  author    = {Lewis, Patrick and Perez, Ethan and Piktus, Aleksandra and Petroni, Fabio and Karpukhin, Vladimir and Goyal, Naman and K{\"u}ttler, Heinrich and Lewis, Mike and Yih, Wen{-}tau and Rockt{\"a}schel, Tim and Riedel, Sebastian and Kiela, Douwe},
  title     = {Retrieval-Augmented Generation for Knowledge-Intensive {NLP} Tasks},
  booktitle = {Advances in Neural Information Processing Systems (NeurIPS)},
  year      = {2020}
}

@inproceedings{ijcai2022p762,
  title     = {A Survey of Vision-Language Pre-Trained Models},
  author    = {Du, Yifan and Liu, Zikang and Li, Junyi and Zhao, Wayne Xin},
  booktitle = {Proceedings of the Thirty-First International Joint Conference on
               Artificial Intelligence, {IJCAI-22}},
  @publisher = {International Joint Conferences on Artificial Intelligence Organization},
  editor    = {Lud De Raedt},
  pages     = {5436--5443},
  year      = {2022},
  month     = {7},
  @address   = {Vienna, Austria},
  note      = {Survey Track},
  @doi       = {10.24963/ijcai.2022/762},
  @url       = {https://doi.org/10.24963/ijcai.2022/762},
}

@inproceedings{devlin-etal-2019-bert,
    title = "{BERT}: Pre-training of Deep Bidirectional Transformers for Language Understanding",
    author = "Devlin, Jacob  and
              Chang, Ming-Wei  and
              Lee, Kenton  and
              Toutanova, Kristina",
    booktitle = "Proceedings of the 2019 Conference of the North {A}merican Chapter of the Association for Computational Linguistics (NAACL): Human Language Technologies, Volume 1 (Long and Short Papers)",
    month = jun,
    year = "2019",
    @address = "Minneapolis, Minnesota",
    @publisher = "Association for Computational Linguistics",
    @url = "https://doi.org/10.18653/v1/N19-1423",
    @doi = "10.18653/v1/N19-1423",
    pages = "4171--4186"
}

@inproceedings{sarto2022retrieval,
  author    = {Sarto, Sara and Cornia, Marcella and Baraldi, Lorenzo and Cucchiara, Rita},
  title     = {Retrieval-Augmented Transformer for Image Captioning},
  booktitle = {CBMI},
  year      = {2022},
  @publisher = {Association for Computing Machinery},
  @address   = {New York, NY, USA},
  pages     = {1--7},
  @doi       = {10.1145/3549555.3549585},
}

@inproceedings{lin2014microsoft,
  author    = {Lin, Tsung{-}Yi and Maire, Michael and Belongie, Serge and Hays, James and Perona, Pietro and Ramanan, Deva and Doll{\'a}r, Piotr and Zitnick, C. Lawrence},
  editor    = {Fleet, David and Pajdla, Tom{\'a}{\v{s}} and Schiele, Bernt and Tuytelaars, Tinne},
  title     = {Microsoft {COCO}: Common Objects in Context},
  booktitle = {Computer Vision -- ECCV 2014},
  series    = {Lecture Notes in Computer Science},
  volume    = {8693},
  pages     = {740--755},
  @publisher = {Springer},
  @address   = {Cham},
  year      = {2014},
  @doi       = {10.1007/978-3-319-10602-1_48}
}

@inproceedings{ramos2023extra,
  author    = {Ramos, Rita and Elliott, Desmond and Martins, Bruno},
  title     = {Retrieval-augmented Image Captioning},
  booktitle = {Proceedings of the 17th Conference of the European Chapter of the Association for Computational Linguistics (EACL)},
  @address   = {Dubrovnik, Croatia},
  @publisher = {Association for Computational Linguistics},
  pages     = {3666--3681},
  year      = {2023},
  month     = may,
  @doi       = {10.18653/v1/2023.eacl-main.266},
}

@inproceedings{li2024evcap,
  author    = {Li, Jiaxuan and Vo, Duc Minh and Sugimoto, Akihiro and Nakayama, Hideki},
  title     = {EVCap: Retrieval-Augmented Image Captioning with External Visual-Name Memory for Open-World Comprehension},
  booktitle = {Proceedings of the IEEE/CVF Conference on Computer Vision and Pattern Recognition (CVPR)},
  year      = {2024},
  pages     = {20086--20096},
  @publisher = {IEEE},
  @address   = {Seattle, WA, USA},
  month     = jun,
}

@misc{li2024understanding,
  author        = {Li, Wenyan and Li, Jiaang and Ramos, Rita and Tang, Raphael and Elliott, Desmond},
  title         = {Understanding Retrieval Robustness for Retrieval-Augmented Image Captioning},
  year          = {2024},
  note          = {Preprint at \url{https://arxiv.org/abs/2406.02265}}
}

@misc{wu2024dir,
  author        = {Wu, Hao and Zhong, Zhihang and Sun, Xiao},
  title         = {DIR: Retrieval-Augmented Image Captioning with Comprehensive Understanding},
  year          = {2024},
  note          = {Preprint at \url{https://arxiv.org/abs/2412.01115}}
}

@inproceedings{ngo2023comprehensive,
  author    = {Ngo, Bach Hoang and Nguyen, Dat Thanh and Do{-}Tran, Nhat{-}Tuong and Pham Huy, Thien Phuc and An, Minh{-}Hung and Nguyen, Tuan{-}Ngoc and Nguyen, Hoang Loi and Dinh, Vinh and Dinh, Vinh},
  title     = {Comprehensive Visual Features and Pseudo Labeling for Robust Natural Language-Based Vehicle Retrieval},
  booktitle = {Proceedings of the IEEE/CVF Conference on Computer Vision and Pattern Recognition (CVPR) Workshops},
  pages     = {5409--5418},
  year      = {2023},
  @publisher = {IEEE},
  @address   = {Vancouver, BC, Canada}
}

@inproceedings{nguyen2023multimodal,
  author    = {Nguyen, Tien{-}Huy and Nguyen{-}Huu, Hoang{-}Long and Le, Thien{-}Doanh and Tran, Huu{-}Loc and Le{-}Tran, Quoc{-}Khanh and Ngo, Hoang{-}Bach and An, Minh{-}Hung and Dinh, Quang{-}Vinh},
  title     = {Multimodal Fusion in NewsImages 2023: Evaluating Translators, Keyphrase Extraction, and {CLIP} Pre-Training},
  booktitle = {Working Notes Proceedings of the MediaEval 2023 Workshop},
  series    = {CEUR Workshop Proceedings},
  volume    = {3658},
  @address   = {Aachen},
  @publisher = {CEUR-WS.org},
  year      = {2024},
  @url       = {https://ceur-ws.org/Vol-3658/paper9.pdf}
}

@inproceedings{nguyen2025openevents,
  author    = {Nguyen, Hieu and Nguyen, Phuc{-}Tan and Tran, Thien{-}Phuc and Nguyen, Minh{-}Quang and Nguyen, Tam V. and Tran, Minh{-}Triet and Le, Trung{-}Nghia},
  title     = {OpenEvents V1: Large-Scale Benchmark Dataset for Multimodal Event Grounding},
  booktitle = {Proceedings of the ACM International Conference on Multimedia (ACM MM)},
  year      = {2025}
}

@inproceedings{eventa25,
  author    = {Tran, Thien{-}Phuc and Nguyen, Minh{-}Quang and Tran, Minh{-}Triet and Nguyen, Tam V. and Do, Trong{-}Le and Ly, Duy{-}Nam and Huynh, Viet{-}Tham and Le, Khanh{-}Duy and Tran, Mai{-}Khiem and Le, Trung{-}Nghia},
  title     = {Event-Enriched Image Analysis Grand Challenge at {ACM} Multimedia 2025},
  booktitle = {Proceedings of the ACM International Conference on Multimedia (ACM MM)},
  year      = {2025}
}

@inproceedings{yang2023semantic,
  author    = {Yang, Shuyu and Zhou, Yinan and Wang, Yaxiong and Wu, Yujiao and Zhu, Li and Zheng, Zhedong},
  title     = {Towards Unified Text-based Person Retrieval: A Large-scale Multi-Attribute and Language Search Benchmark},
  booktitle = {Proceedings of the 31st ACM International Conference on Multimedia (MM '23)},
  year      = {2023},
  @publisher = {Association for Computing Machinery},
  @address   = {New York, NY, USA},
}

@misc{muennighoff2024m3,
  author        = {Muennighoff, Niklas and Autry, Luke and Wang, Qifan and Neyshabur, Behnam and Rajani, Nazneen and Ren, Xiang},
  title         = {M3-Embedding: A Purely Text-based Embedding Model for Multilingual, Multi-task Retrieval},
  year          = {2024},
  eprint        = {2402.03216},
  archivePrefix = {arXiv},
  primaryClass  = {cs.CL},
  @url           = {https://arxiv.org/abs/2402.03216}
}

@inproceedings{hessel2021clipscore,
  author    = {Hessel, Jack and Holtzman, Ari and Forbes, Maxwell and Choi, Yejin},
  title     = {CLIPScore: A Reference-free Evaluation Metric for Image Captioning},
  booktitle = {Proceedings of the 2021 Conference on Empirical Methods in Natural Language Processing (EMNLP)},
  year      = {2021},
  pages     = {7514--7528},
  @publisher = {Association for Computational Linguistics},
  @doi       = {10.18653/v1/2021.emnlp-main.599},
}

@inproceedings{ji2023multimodal,
  author    = {Ji, Wei and Wei, Yinwei and Zheng, Zhedong and Fei, Hao and Chua, Tat-Seng},
  title     = {Deep Multimodal Learning for Information Retrieval},
  booktitle = {Proceedings of the 31st ACM International Conference on Multimedia (MM '23)},
  pages     = {9739--9741},
  year      = {2023},
  @publisher = {Association for Computing Machinery},
  @address   = {New York, NY, USA},
  @doi       = {10.1145/3581783.3610949},
  @url       = {https://dl.acm.org/doi/10.1145/3581783.3610949}
}

@inproceedings{radford2021learning,
  author    = {Radford, Alec and Kim, Jong Wook and Hallacy, Chris and Ramesh, Aditya and Goh, Gabriel and Agarwal, Sandhini and Sastry, Girish and Askell, Amanda and Mishkin, Pamela and Clark, Jack and others},
  title     = {Learning Transferable Visual Models From Natural Language Supervision},
  booktitle = {International Conference on Machine Learning (ICML)},
  pages     = {8748--8763},
  year      = {2021},
  organization={PMLR}
}

\end{document}